\documentclass[11pt,a4paper]{article}
\usepackage[hyperref]{acl2021}
\usepackage{times}
\usepackage{latexsym}

\usepackage{microtype}

\aclfinalcopy 
\def\aclpaperid{768} 

\usepackage{amstext}
\usepackage{booktabs}
\usepackage{graphicx}
\usepackage{amsfonts}
\usepackage{bm}
\usepackage{subfigure}
\usepackage{multirow}
\usepackage{adjustbox}
\usepackage{array}
\usepackage{makecell}
\usepackage{color}

\definecolor{orange}{RGB}{197,90,17}
\definecolor{blue}{RGB}{46,117,182}
\title{Dynamic Semantic Graph Construction and Reasoning for\\ Explainable Multi-hop Science Question Answering\thanks{\hspace{2mm}The work described in this paper is substantially supported by a grant from the Research Grant Council of the Hong Kong Special Administrative Region, China (Project Code: 14204418).}}

\author{
Weiwen Xu, Huihui Zhang, Deng Cai and Wai Lam \\
The Chinese University of Hong Kong \\

{\tt \{wwxu,hhzhang,wlam\}@se.cuhk.edu.hk} \\
{\tt thisisjcykcd@gmail.com}
}

\date{}

\begin{document}
\maketitle
\begin{abstract}

Knowledge retrieval and reasoning are two key stages in multi-hop question answering (QA) at web scale. Existing approaches suffer from low confidence when retrieving evidence facts to fill the knowledge gap and lack transparent reasoning process. In this paper, we propose a new framework to exploit more valid facts while obtaining explainability for multi-hop QA by dynamically constructing a semantic graph and reasoning over it. We employ Abstract Meaning Representation (AMR) as semantic graph representation. Our framework contains three new ideas: (a) {\tt AMR-SG}, an AMR-based Semantic Graph, constructed by candidate fact AMRs to uncover any hop relations among question, answer and multiple facts. (b) A novel path-based fact analytics approach exploiting {\tt AMR-SG} to extract active facts from a large fact pool to answer questions. (c) A fact-level relation modeling leveraging graph convolution network (GCN) to guide the reasoning process. Results on two scientific multi-hop QA datasets show that we can surpass recent approaches including those using additional knowledge graphs while maintaining high explainability on OpenBookQA and achieve a new state-of-the-art result on ARC-Challenge in a computationally practicable setting.
\end{abstract}
\section{Introduction}

Multi-hop QA is one of the most challenging tasks that benefits from explainability as it mimics the human question answering setting, where multi-hop QA requires both the collection of information from large external knowledge resources and the aggregation of retrieved facts to answer complex natural language questions~\cite{yang-etal-2018-hotpotqa}.
\begin{figure}[ht]
    \centering
\includegraphics[scale=0.45]{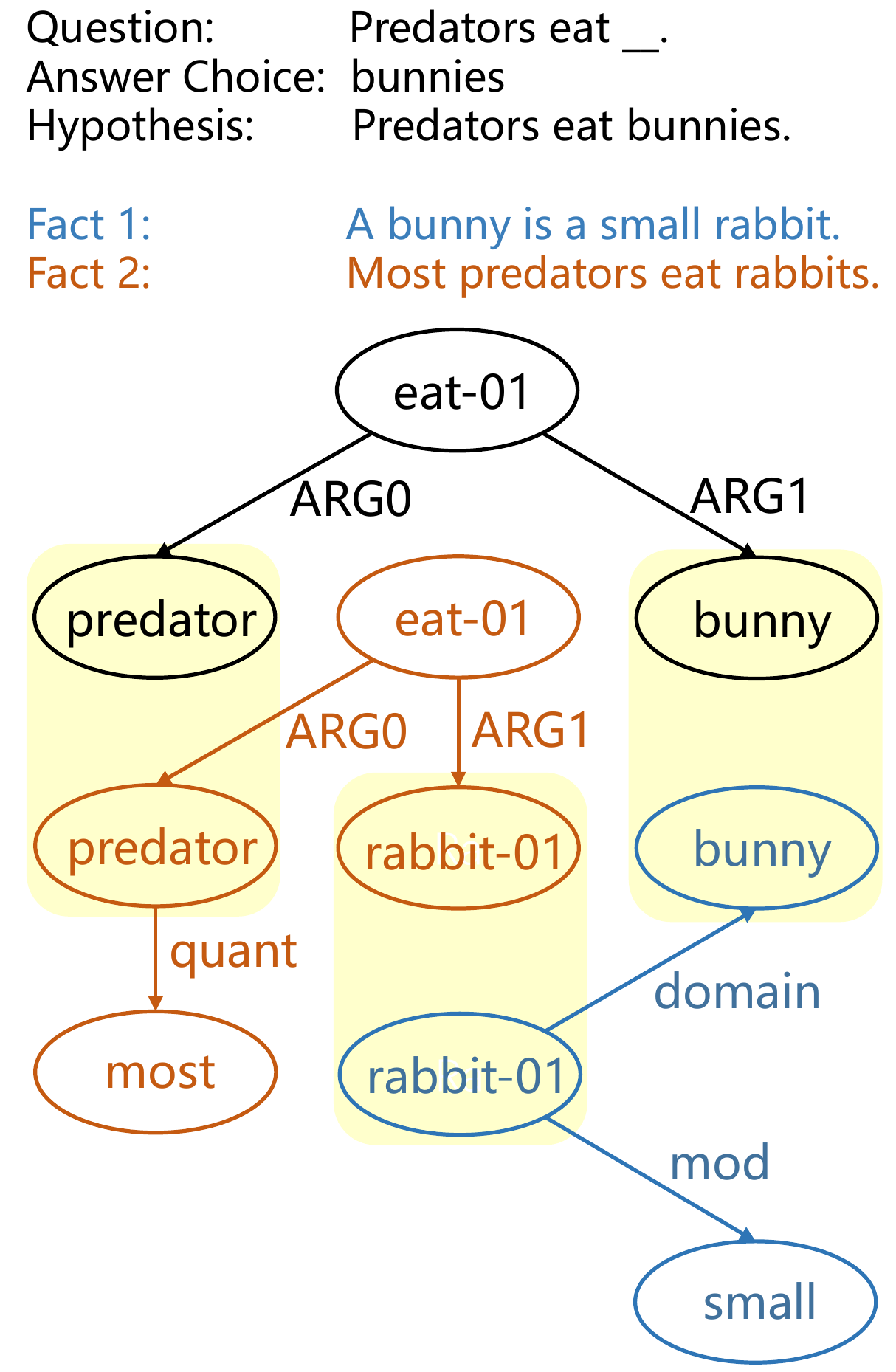} 
       \caption{The AMR of the hypothesis (black), \textcolor{blue}{Fact 1} and \textcolor{orange}{Fact 2}. A hypothesis is a statement derived from a question and a choice. The hypothesis AMR can be inferred by relevant fact AMRs.}
    \label{fig:amr}
\end{figure}

Currently, external knowledge is mostly stored in two forms -- textual and graph structure (e.g. Knowledge Graph (KG)). Textual corpora contain rich and diverse evidence facts, which are ideal knowledge resources for multi-hop QA. Especially with the success of pretrained models~\cite{devlin-etal-2019-bert, liu2019roberta, lan2019albert}, we can get powerful representations for such textual facts. However, retrieving relevant and useful facts to fill the knowledge gap for inferring the answer is still a challenging problem. In addition, the reasoning process over the facts is hidden by the unexplainable neural network, which hinders the deployment of real-life applications. On the other hand, KG is able to provide structural clues about relevant entities for explainable predictions~\cite{feng-etal-2020-scalable,saxena-etal-2020-improving, xu2020fusing}. But it is known to suffer from sparsity, where complex question clues are unlikely to be covered by the closed-form relations in KG~\cite{zhao2020complex,zhang2020gmh}. Another issue is that KG requires large human labor and is easy to become outdated if not maintained timely. 

To take advantages of both rich textual corpora and explicit graph structure and make it compatible to all textual knowledge, we explore the usefulness of Abstract Meaning Representation (AMR) as a graph annotation to a textual fact. AMR~\cite{banarescu-etal-2013-abstract} is a semantic formalism that represents the meaning of a sentence into a rooted, directed graph. Figure~\ref{fig:amr} shows some examples of AMR graphs, where nodes represent concepts and edges represent the relations. Unlike other semantic role labeling that only considers the relations between predicates and their arguments~\cite{song-etal-2019-semantic}, the aim of AMR is to capture every meaningful content in high-level abstraction while removing away inflections and function words in a sentence. As a result, AMR allows us to explore textual facts and simultaneously attributes them with explicit graph structure for explainable fact quality assessment and reasoning.

In this paper, we propose a novel framework that incorporates AMR to make explainable knowledge retrieval and reasoning for multi-hop QA. Our framework works on textual knowledge, which is easy to obtain and allows us to get informative facts. The introduced AMR serves as a bridge that enables an explicit reasoning process over a graph structure among questions, answers and relevant facts. As exemplified in Figure~\ref{fig:amr}, a hypothesis is first derived from a question and an answer choice. We then parse the hypothesis and a large number of facts to corresponding AMRs. After that, we dynamically construct {\tt AMR-SG} for each question-choice pair by merging the AMRs of its hypothesis and relevant facts. Unlike previous works on multi-hop QA that rely on existing KGs to find relations among entities~\cite{wang-etal-2020-connecting, feng-etal-2020-scalable}, our proposed {\tt AMR-SG} is dynamically constructed, which reveals intrinsic relations of facts and can naturally form any-hop connections. After construction, we analyze all connected paths starting from the question to the answer on {\tt AMR-SG}. We focus the consideration of facts on those paths because they together connect the question with the answer, indicating their active roles in filling the knowledge gap. The connections of facts on {\tt AMR-SG} can be further used as the supervision for downstream reasoning. Therefore, we adopt GCN~\cite{Kipf:2016tc} to model the fact-level information passing. 

Experimental results demonstrate that our approach outperforms previous approaches that use additional KGs. It obtains 81.6 accuracy on OpenBookQA~\cite{OpenBookQA2018}, and pushes the state-of-the-art result on ARC-Challenge~\cite{clark2018think} to 68.94 in a computationally practicable setting.
\section{Related Work}
\paragraph{Multi-hop QA with External Resource.}
Despite the success of pretrained model in most Natural Language Processing (NLP) tasks, it performs poorly in multi-hop QA, where some information is missing to answer questions~\cite{zhu2021retrieving}.

\begin{figure*}[ht]
    \centering
\includegraphics[scale=0.49]{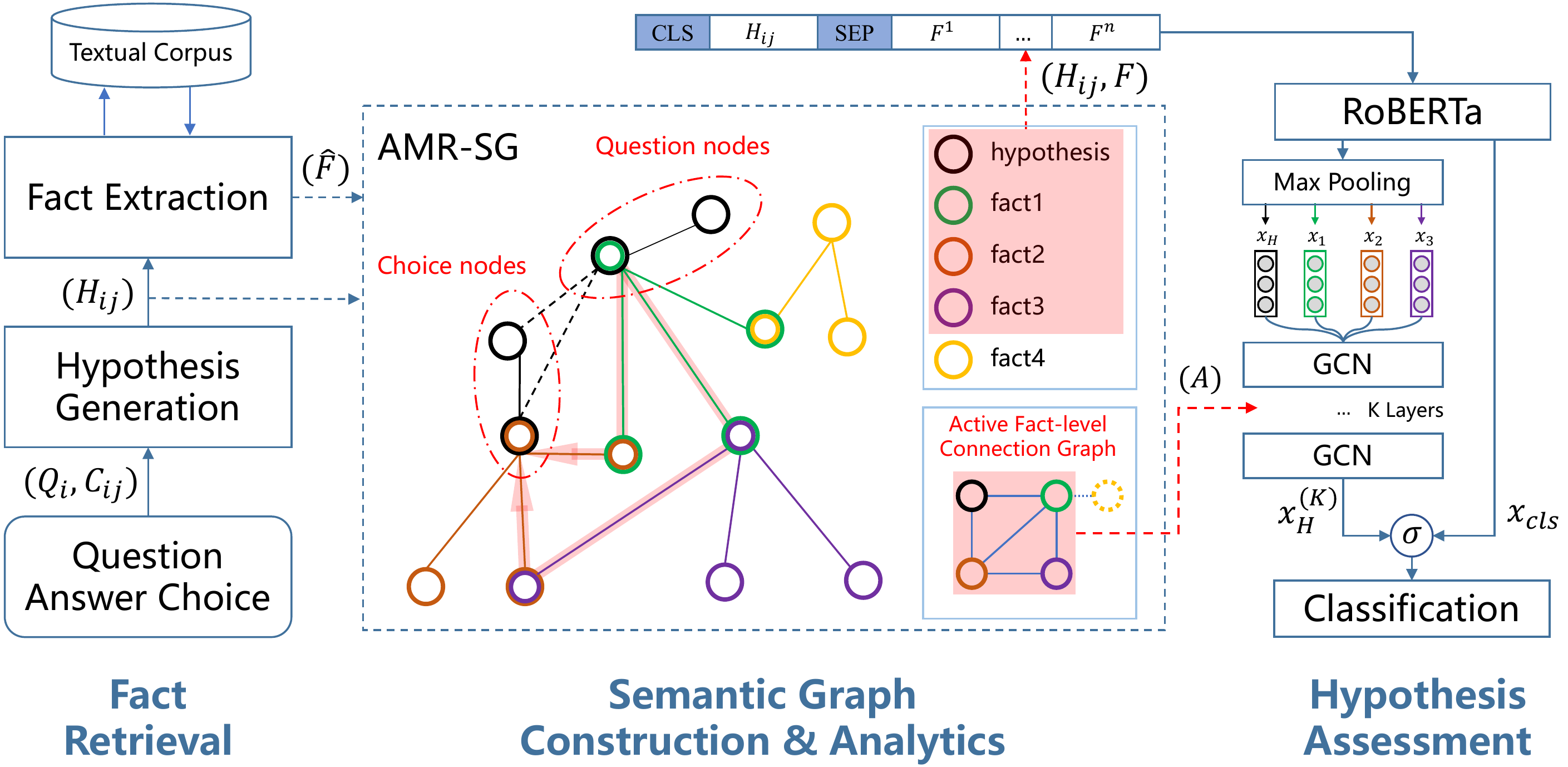} 
       \caption{Overall architecture of our proposed model. The black dash lines in {\tt AMR-SG} indicate that we cut the connection between question nodes and choice nodes. The pink arrows indicate two paths that can be spotted in {\tt AMR-SG}. Facts with red background are active facts detected. The dashed node \textit{Active Fact-level Connection Graph} indicates fact4 is not considered as a valid node as it is not an active fact.}
    \label{fig:model}
\end{figure*}

Textual corpora contain rich and diverse knowledge, which is likely to cover the clues to answer complex questions. \citet{banerjee-etal-2019-careful} demonstrate some carefully designed queries can effectively retrieve relevant facts. \citet{yadav-etal-2019-quick,DBLP:conf/emnlp/DengZL20} extract groups of evidence facts considering the relevance, overlap and coverage, but such method requires exponential computation in the retrieval step. \citet{feldman-el-yaniv-2019-multi, yadav-etal-2020-unsupervised} construct a fact chain by iteratively reformulating the query to focus on the missing information. However, the fact chain often grows obliquely as a result of the failure of first fact retrieval, making the QA model brittle. As some recent QA datasets~\cite{yang-etal-2018-hotpotqa, OpenBookQA2018, khot2020qasc} annotate a gold evidence fact for each question, it enables training supervised classifier to identify the correct fact driven by a query~\cite{nie-etal-2019-revealing, qiu-etal-2019-dynamically, tu2020select, banerjee2020knowledge}. \citet{min-etal-2018-efficient} take a further step to jointly predict the answer span and select evidence facts in a unified model. Though these supervised retrievers have achieved impressive improvement, they heavily rely on the annotated gold facts, which are not always available in real-world applications.

In addition, previous works also explore the effectiveness of structured knowledge by either encoding the nodes~\cite{yang-mitchell-2017-leveraging, wang2019improving}, triples~\cite{mihaylov-frank-2018-knowledgeable, wang-etal-2020-connecting}, paths~\cite{lin-etal-2019-kagnet, Lei_2020} or tabular~\cite{zhu2021tatqa} to capture the missing information. Other works avoid the sparsity of KGs by constructing KGs directly from textual knowledge. OpenIE~\cite{saha-mausam-2018-open} is widely used in knowledge base question answering to extract entity-relation triples~\cite{bosselut-etal-2019-comet, zhao2020complex,DBLP:conf/aaai/DengXLYDFLS19}. However, OpenIE favors precision over recall, which is not necessarily effective to form connections among diverse evidence facts for multi-hop QA. Wikipedia  contains internal hyperlinks, which are effective to build graph connections from unstructured articles~\cite{Asai2020Learning, liu2020rikinet}. However, such hyperlinks are not available in most textual corpora.

\paragraph{AMR.}
Recent success in AMR research makes it possible to benefit downstream tasks, such as summarization~\cite{takase-etal-2016-neural, dohare2017text, liao-etal-2018-abstract}, event detection~\cite{li-etal-2015-improving-event} and machine translation~\cite{song-etal-2019-semantic}. In the domain of QA, AMR has been used to form logic queries and conduct symbolic reasoning~\cite{mitra2016addressing, kapanipathi2020question}. Comparing to name entity~\cite{zhong-etal-2020-reasoning} or other cross-sentence annotations~\cite{lei2018linguistic,DBLP:conf/sigir/ZhangDL20}, we use AMR to build our semantic graph because it is align-free and can be easily adapted to powerful pretrained models.
\section{Framework Description}
In this paper, we consider the multi-hop QA in the form of multi-choice, where a question $Q_i$ is provided with $J$ answer choices $C_{ij}, j \in \{1,2,...,J\}$. As shown in Figure~\ref{fig:model}, our framework consists of three components: (1) a Fact Retrieval component to retrieve evidence facts $\hat{F}=\{\hat{F}^1,...,\hat{F}^m\}$\footnote{We omit the subscript $ij$ for simplicity.} for each question-choice pair from a large textual corpus; (2) a Semantic Graph Construction \& Analytics component that dynamically constructs a semantic graph, named {\tt AMR-SG}, to select active facts  $F=\{F^1,...,F^n\}$ from $\hat{F}$ and capture their relations $A$; and (3) a Hypothesis Assessment component that classifies whether the question-choice is correct, given the active facts and their relations in (2).

\subsection{Fact Retrieval}
\paragraph{Hypothesis Generation.}
As shown in Figure~\ref{fig:model}, we first generate a hypothesis $H_{ij}$ for the $i^{th}$ question and the $j^{th}$ choice. A hypothesis is a completed statement derived from each question-choice pair. Comparing to simply concatenating the question and the choice, a hypothesis contains less meaningless words and maintain a good grammatical structure, which can avoid retrieving noisy facts and allow AMR parser to generate high-quality AMR graphs. We generate hypotheses by the rule-based model of~\citet{demszky2018transforming}. For some unsolvable cases, we directly concatenate the question and the choice. We apply this process for all training, develop and test sets.

\paragraph{Fact Extraction.}
We retrieve a pool of evidence facts $\hat{F}$ for each hypothesis separately using \textit{Elasticsearch}~\cite{gormley2015elasticsearch}. We set a large size $m$ of the fact pool to cover as many valid facts as possible. 
\subsection{Semantic Graph Construction \& Analytics}
\textit{Active facts} $F$ are facts that really fill the knowledge gap between question and choice. The activeness of a fact cannot be simply determined by comparing it with the hypothesis, as multi-hop QA requires multiple facts to complete the reasoning chain. Therefore, we need to filter out facts that are just partially related and focus on the consideration of active facts and their roles in the reasoning chain. In this component, we first construct {\tt AMR-SG}. Then, we propose a path-based analytics approach to extract active facts and construct an \textit{Active Fact-level Connection Graph} to capture their relations with the question and the answer choice.

\subsubsection{AMR-SG Construction}
As the nodes of AMR are high-level abstraction of concepts conveyed in the corresponding textual fact, two AMRs sharing the same node indicate that they concern about the same concept, which shows their correlation. This motivates us to construct {\tt AMR-SG}, shown in Figure~\ref{fig:model}, to represent the relations of the corresponding hypothesis and evidence facts for each question-choice pair.

We leverage the state-of-the-art AMR parser~\cite{cai-lam-2020-amr} to generate AMR $G=\{G^H, G^1,...,G^m\}$ for a hypothesis and all facts in the corresponding fact pool, where $G^H$, $G^i$ are the AMR of the hypothesis and the $i^{th}$ fact respectively. AMR is also a directed and edge-labeled graph, which implies information specified in the edge is propagated in one pre-defined direction. However, such inner-AMR (edge labels and directions) information does not contribute to inter-AMR relations. Therefore, we only care about if there exists an edge between two nodes but ignore the edge labels and directions.

During construction, we regard $G^H$ as the start point of {\tt AMR-SG}. Then, we incrementally find one fact AMR in the fact pool sharing some nodes with it and add this fact AMR onto it by merging the shared nodes. The merging operation stops when no AMR can be added onto {\tt AMR-SG} or the fact pool is empty. In fact, as shown in Figure~\ref{fig:model}, we do not change the architecture of each individual AMR, but reuse some shared nodes as the nodes in {\tt AMR-SG}. Note that, some nodes are over-general, which are not appropriate to connect two AMRs (e.g. {\tt  (p/planet :name(n/name :op1"Earth"))}, the node {\tt n/name} is an over-general concept). Fortunately, such over-general nodes always have non-node attributes (e.g. {\tt Earth} of {\tt n/name}) that shows the specific referent. Therefore, we replace the nodes with their non-node attributes if any to address this issue.

\subsubsection{Path-based Analytics}
Current multi-hop QA models are hindered by the quality of retrieved facts~\cite{banerjee-etal-2019-careful}. We address this issue by a path-based analytics approach to guarantee the selected facts having a positive effect to answer the question.

As shown in Figure~\ref{fig:model}, {\tt AMR-SG} reveals any-hop relations of the hypothesis and all facts. Completed paths can be spotted out of $G^H$ to connect the question nodes with the choice nodes by passing through multiple facts. These facts, which together provide the missing knowledge to maintain complete reasoning chains, are active facts that we want to extract.

Specifically, we split the nodes of $G^H$ into \textit{question nodes} $Q^H$ and \textit{choice nodes} $C^H$. Question nodes represent the concepts extracted in the question text. As one question is provided with $J$ choices, where we can generate $J$ hypothesis AMRs. We take the shared nodes of these AMRs as  $Q^H$, while the remaining as $C^H$:
\begin{equation}
    Q^H_{ij} = \cap_{j=1}^J\{v| v \in G^H_{ij}\}
\end{equation}
\begin{equation}
        C^H_{ij} = \{v| v \in G^H_{ij}, v \notin Q^H_{ij}\}, j=1,...,J
\end{equation}

We cut the edges between $Q^H$ and $C^H$ to guarantee the paths are spotted outside $G^H$. Then we apply depth-first search on {\tt AMR-SG} to find all paths that connect at least one question node and one choice node, including the path that does not have a minimum length (e.g. the path passing through fact3 in Figure~\ref{fig:model}). All facts that the paths pass through (one node in and another node out) are considered as active facts. This is because we try to cover more facts as long as they do not deviate from the correct reasoning direction to provide enough information for QA model.

In addition, the any-hop relations of the hypothesis and active facts in {\tt AMR-SG} can be used for a hypothesis to precisely aggregate knowledge from relevant facts to reduce ambiguity during the reasoning process. Therefore, we construct an \textit{Active Fact-level Connection Graph} from {\tt AMR-SG} to capture such relations among the hypothesis and all active facts. As shown in Figure~\ref{fig:model}, each node in \textit{Active Fact-level Connection Graph} is either the hypothesis or an active fact. We draw an edge between two facts (include hypothesis) if they share one concept node in {\tt AMR-SG}. 

\subsection{Hypothesis Assessment with Fact-level Reasoning}
As shown in Figure~\ref{fig:model}, we concatenate the hypothesis with all active facts, where {\tt [SEP]} token is inserted between the two texts and {\tt [CLS]} is put at the beginning of the sequence. We feed the whole sequence into a pretrained model based on RoBERTa~\cite{liu2019roberta} architecture to get the hidden representation of each token.

Then, \textit{Active Fact-level Connection Graph} is used as an additional supervision in fact-level modeling to guide the reasoning process. Formally, let  $s^{H}_{1:{l_H}} \in \mathbb{R}^{l_H \times d}$, $s^{i}_{1:{l_i}} \in \mathbb{R}^{l_i \times d}$ be the hidden representations of the hypothesis and the $i^{th}$ active fact respectively, where $l_H$, $l_i$ denote the length and $d$ is the dimension of the representation. A max pooling layer is applied over these hidden representations to get the node representations respectively:
\begin{equation}
\begin{array}{l}
    x_H = \textbf{MaxPool}(s^{H}_{1:{l_H}}) \in \mathbb{R}^{1 \times d}\\
         x_i = \textbf{MaxPool}(s^{i}_{1:{l_i}}) \in \mathbb{R}^{1 \times d}, i=1,...,n
\end{array}
\end{equation}
The connections of hypothesis ($0^{th}$) and active facts in \textit{Active Fact-level Connection Graph} can be viewed as an adjacency matrix $A \in \mathbb{R}^{(n+1) \times (n+1)}$, where
\begin{equation}
    A_{ij}=\left\{
\begin{array}{ll}
     1& \text{if }  F^i \text{ is connected with } F^j\\
     0& \text{otherwise}
\end{array}
\right.
\end{equation}
As there is no edge information in the graph, a simple GCN is enough to model the knowledge fusion among the hypothesis and multiple active facts in the reasoning process. We also introduce multi-head mechanism~\cite{NIPS2017_3f5ee243} to stabilize the learning of different knowledge:
\begin{equation}
    X^{(k)}=[head_1^{(k)}:...:head_h^{(k)}] 
\end{equation}
where $[:]$ denotes concatenation operation, $X^{(k)}$ is the node states at the $k^{th}$ layer, $X^{(0)}=[x_H;x_1;...;x_n]$, $[;]$ denotes the sequential concatenation operation, $head_i$ is the $i^{th}$ head. Specifically, we compute the nodes states by aggregating knowledge from their neighboring nodes in each layer:
\begin{equation}
    head_i^{(k)}=\mathrm{ReLU}(\Lambda X^{(k-1)}W_i^{(k)}) 
\end{equation}
where $W_i^{(k)}\in \mathbb{R}^{d \times (d/h)}$ is the projection matrix of $head_i$ at the $k^{th}$ layer, $h$ is the head number. $\Lambda$ is the normalization constant to avoid scale changing:
\begin{equation}
\begin{array}{l}
    \Lambda=D^{-1/2}AD^{-1/2}\\
         D_{ii}=\sum_jA_{ij}
\end{array}
\end{equation}
After that, a $\sigma$ gate is applied to calculate how much knowledge can be propagated to score the question-choice pair:
\begin{equation}
    \lambda = \sigma( W^{\lambda}[x_{cls}:x^{(K)}_{H}]+b^{\lambda})
\end{equation}
\begin{equation}
    s(q,a) = W^o(\lambda x^{(K)}_{H} +(1-\lambda)x_{cls}) + b^o
\end{equation}
where $W^{\lambda} \in \mathbb{R}^{1 \times 2d}$, $W^o \in \mathbb{R}^{d \times d}$, $b^{\lambda}$, $b^o$ are the parameters. We get the final probability by normalize all question-choice pairs with softmax.
\section{Experiments}
\subsection{Datasets}
We evaluate our approach on two multi-choice multi-hop QA datasets: ARC-Challenge~\cite{clark2018think} and OpenBookQA~\cite{OpenBookQA2018}. The textual corpus we use for both datsets is ARC Corpus~\cite{clark2018think}, which contains about 14M science facts.
OpenBookQA and ARC-Challenge have their leaderboards with train, develop and test sets publicly available. 
we follow~\citet{AristoRoBERTAv7} to combine the training set of  OpenBookQA (4957), ARC-Easy (2251), ARC-Challenge (1119) and RegLivEnv (665) as the final training set of ARC-Challenge task. The data splits
is shown in Table~\ref{tab:dataset}.
\begin{table}[t]
    \centering
    \begin{tabular}{cccc}
    \toprule
         & Train & Dev & Test\\ \hline
         OpenBookQA& 4957 &500 &500\\
         ARC-Challenge& 8992 &299 &1172\\ \bottomrule
    \end{tabular}
    \caption{Number of instances in each dataset.}
    \label{tab:dataset}
\end{table}

For ARC-Challenge, we retrieve 100 facts to form the fact pool. Based on this, we select up to 20 active facts using our approach as the context for each question-choice pair.\footnote{We can only reproduce the results similar to~\citet{AristoRoBERTAv7} using 20 facts as the context.} OpenBookQA provides an accompanying open-book of 1326 science facts, which are highly related to the questions in this dataset. Therefore, for OpenBookQA, we retrieve 10 facts from the open-book and another 90 facts from ARC Corpus, forming the 100 facts in the fact pool. We then select up to 15 active facts using our approach as the context.

\subsection{Implementation}
We implement our approach on two pretrained models: RoBERTa~\cite{liu2019roberta} and AristoRoBERTa~\cite{AristoRoBERTAv7}. AristoRoBERTa employs the RoBERTa architecture but uses RACE~\cite{lai-etal-2017-race} to first fine-tune the RoBERTa model. We prepare active facts as the context to further fine-tune the model with the target dataset.
For OpenBookQA, we continue to fine-tune the QA model following the same procedure as \citet{AristoRoBERTAv7}, where the initial learning rate is 2e-5, the batch size is 12 and the max sequence length is 256. For ARC-Challenge, the initial learning rate, the batch size and the max sequence length are 1e-5, 6, and 416 respectively. We use grid search to find optimal hyper-parameters, where the learning rate is chosen from \{5e-6, 1e-5, 2e-5\}, the batch size is chosen from \{4, 6, 8, 12, 16\}. The number of GCN layer $K$ is chosen from \{1,\textbf{2},3,4\}, while the head number $h$ is the RoBERTa-Large default value.\footnote{Our code is available at: \url{https://github.com/wwxu21/AMR-SG}}

We introduce 6M parameters of the fact-level reasoning module in addition to 355M of RoBERTa-Large. We run all experiments on one TITAN RTX card, which takes about 1 hour and 3 hours to complete the training of OpenBookQA and ARC-Challenge respectively.

\subsection{Comparison Methods}
We compare with recent existing methods that make use of similar power of pretrained models in order to conduct a fair comparison. These include the baseline AristoRoBERTaV7~\cite{AristoRoBERTAv7} finetuned on top of AristoRoBERTa, KF-SIR~\cite{banerjee2020knowledge} that exploits the knowledge fusion among facts, FreeLB~\cite{Zhu2020FreeLB:} that tackles the robustness issue and another three methods leveraging an additional knowledge graph~\cite{speer2017conceptnet} in addition to the textual knowledge: PG~\cite{wang-etal-2020-connecting}, MHGRN~\cite{feng-etal-2020-scalable}, AlBERT + KB.  PG(albert + gpt2, roberta + gpt2) are two implementations with different pretrained model architectures~\cite{liu2019roberta, lan2019albert, radford2019language}, where the latter is more fair to compare with us.

\section{Results}
\subsection{Main Results}
\begin{table}[]
    \centering
    \small
    \begin{tabular}{lp{18mm}<{\centering}p{12mm}<{\centering}p{6mm}}
    \toprule
      \normalsize{Methods} &  Model Architecture &  Additional KG & Test Acc.  \\\midrule
        PG& {\small albert + gpt2}& \checkmark &81.8\\
        PG& {\small roberta + gpt2}&\checkmark &80.2\\
         AlBERT + KB& {\small albert} &\checkmark & 81.0\\
         MHGRN& {\small roberta}&\checkmark &80.6\\
         KF-SIR& {\small roberta} & $\times$ &80.0\\ \midrule
          AristoRoBERTaV7& {\small roberta}& $\times$ & 77.8\\ 
         + AMR-SG-Full & {\small roberta}& $\times$ & \textbf{81.6}\\ \bottomrule
    \end{tabular}
    \caption{Test accuracy on OpenBookQA. Methods using additional KG are ticked.}
    \label{tab:leadobqa}
\end{table}
\paragraph{OpenBookQA.}
The test set accuracy is shown in Table~\ref{tab:leadobqa}. AMR-SG-Full is our full model based on AristoRoBERTa. Results show that AMR-SG-Full can surpass models leveraging additional KG. It demonstrates that the fundamental improvement of AMR-SG-Full comes from the knowledge mining of the textual corpus. However, such knowledge resource has not been fully investigated by existing methods and contains richer and more diverse evidence facts than KGs.  We do not compare with UnifiedQA~\cite{khashabi-etal-2020-unifiedqa} and T5 3B~\cite{raffel2020exploring} as they rely on extremely large pretrained models (at least 3B parameters), which are not fair for comparison. 
\paragraph{ARC-Challenge.}
We also implement AMR-SG-Full on another difficult multi-hop QA dataset: ARC-Challenge. It consists of the questions only answered incorrectly by both a retrieval-based algorithm and a word co-occurrence algorithm~\cite{clark2018think}, which theoretically is not friendly to our approach. As shown in Table~\ref{tab:arc}, we can still obtain 2.47 accuracy improvement comparing to AristoRoBERTaV7 and achieve a new state-of-the-art performance in a computationally practicable setting.

\begin{table}[]
    \centering
    \small
    \begin{tabular}{lc}
    \toprule
      Methods & Test Acc.  \\
 \midrule
        FreeLB~\cite{Zhu2020FreeLB:} & 67.75\\
        arcRoberta  & 67.15\\ 
        xlnet+Roberta & 67.06\\ \midrule
        AristoRoBERTaV7~\cite{AristoRoBERTAv7} & 66.47\\ 
        + AMR-SG-Full & \textbf{68.94}\\\bottomrule
    \end{tabular}
    \caption{Test accuracy on ARC-Challenge. All models use RoBERTa architecture for the pretrained model and do not leverage additional KG.}
    \label{tab:arc}
\end{table}

\subsection{Ablation Study}
\begin{table}[]
    \centering
    \begin{tabular}{lll}
    \toprule
      Methods & Dev & Test \\
\hline \multicolumn{3}{c}{ \small RoBERTa}\\ \hline
    No Fact & 66.8& 64.8 \\
        + Fact Context & 68.2& 68.8 (+4.0)\\
    + Fact Analytics  & 73.2& 73.0 (+4.2)\\
         + Fact-level Reasoning & 72.8& 74.2 (+1.2)\\
         \hline \multicolumn{3}{c}{ \small AristoRoBERTa}\\ \hline
        No Fact & 71.0& 70.0 \\ 
        + Fact Context $\spadesuit$& 78.2& 78.4 (+8.4) \\
         + Fact Analytics& 79.4& 81.4 (+3.0)\\
         + Fact-level Reasoning & \textbf{79.6}& \textbf{81.6} (+0.2)\\ \bottomrule
    \end{tabular}
    \caption{Ablation study of model components on OpenBookQA (adding one component incrementally). $\spadesuit$ is our reimplementation of ~\cite{AristoRoBERTAv7}.}
    \label{tab:ablation}
\end{table}
\begin{table*}[]
    \centering
    \begin{tabular}{lcccccc}
    \toprule
         Facts Composition & Core Fact & \multicolumn{3}{c}{Human Evaluation} & \multicolumn{2}{c}{Test set Accuracy}\\ 
         \cmidrule(lr){3-5}\cmidrule(lr){6-7}
               (total 15 facts)  &Retrieval Accuracy& Rel. & Info. & Comp. & RoBERTa & AristoRoBERTa \\ \midrule
    IR (5/10) & 56.4 & 5.86 & 2.50 & 0.46 & 68.8 & 78.4 \\
    IR (10/5) & 63.6 & 5.20 & 2.24 & 0.42 & 70.4 & 77.4 \\
    IR (15/0) & 68.4 & 3.36 & 1.62 & 0.26 & 72.2 & 77.4 \\
    \midrule
    {\tt AMR-SG} (10/30) & 61.0 & 5.85 & 2.58 & 0.48 & 72.4 & 80.4 \\
    {\tt AMR-SG} (10/100) & 61.0 & 6.22 & 2.98 & 0.56 & \textbf{74.2} & \textbf{81.6} \\
    \bottomrule
    \end{tabular}
    \caption{Automatic and Human Evaluation of the evidence facts on OpenBookQA. IR (x/y) indicates we use simple IR system to retrieve x core facts and y common facts. {\tt AMR-SG} (x/y) indicates we construct {\tt AMR-SG} with x core facts and y common facts, based on which we then select 15 active facts and extract their relations. }
    \label{tab:he}
\end{table*}

We conduct ablation study by incrementally adding each component of AMR-SG-Full to investigate its effectiveness on two pretrained models in Table~\ref{tab:ablation}. We include the analysis on RoBERTa because it is a more general and widely used pretrained model. 

We start from the vanilla pretrained models, where no textual facts are provided (denoted as \textit{No Fact}). We retrieve 15 facts as the context to create the first variant (denoted as \textit{+ Fact Context}). The purpose is to test the contribution of the facts retrieved by the simple information retrieval (IR) system (\textit{Elasticsearch}). We continue to add the path-based fact analytics component (denoted as \textit{+ Fact Analytics}). In fact, this variant merely use the facts selected from {\tt AMR-SG} to fine-tune the pretrained models. On top of both two pretrained models, we observe a great performance improvement, where the improvement brought by \textit{+ Fact Analytics} is higher than \textit{+ Fact Context} on top of RoBERTa, which demonstrates this component can effectively select useful facts to fill the knowledge gap that have not been covered by the IR system. We finally equip our model with the fact-level reasoning component (denoted as \textit{+ Fact-level Reasoning}). From the results, we can observe that this component performs well on top of RoBERTa, but has very little effect on top of AristoRoBERTa. This is because this component tries to infuse some fact-level connections to ease the reasoning process of the model. Such information can be learned automatically by the model itself if exposed to enough in-domain data (AristoRoBERTa). Nevertheless, the fact-level reasoning is a more general method when such data is unavailable. 

\section{Explainability Analysis}
\subsection{Analysis of AMR-SG}
\paragraph{Impact of Evidence Facts.}
As discussed above, the major improvement of our approach comes from more useful facts selected for each question-choice pair. In this section, we take a deep look at the quality and the composition of those facts on OpenBookQA. We derive five variants by varying the composition of core (facts retrieved from open-book) or common (facts from ARC Corpus) facts. For core facts, as open-book annotates one gold core fact for each question, the retrieval accuracy of the gold fact is a natural way to evaluate the quality. For common facts, we conduct human analysis to evaluate the quality from three aspects: (1) Relatedness: Does the retrieved fact related to the question or the answer? (2) Informativeness: Does the retrieved fact provided useful information to answer the question? (3) Completeness: Do all retrieved facts together fill the knowledge gap to completely answer the question? We randomly sample 50 questions and evaluate the evidence facts corresponding to the correct answer choice, where one fact would contribute 1 score if it meets the requirement of Relatedness or Informativeness respectively and all 15 facts contribute 1 score if they together meet the requirement of Completeness. Evaluation results are presented in Table~\ref{tab:he}.

When varying the fact composition of IR variants, we find the gold core fact retrieval accuracy has a positive impact on the final accuracy on top of RoBERTa. At this stage, some questions can be inferred sufficiently with the gold core facts. Higher retrieval accuracy accounts for more questions of this kind to be correctly answered. However, this advantage is not as obvious for AristoRoBERTa. Our human evaluation reveals that such facts are unlikely to form a complete reasoning chain, making it hard for real multi-hop reasoning.

On the other hand, our approach directly models the intrinsic fact relations, where the path-based analytics ensures that the facts selected are in the reasoning chain from the question to the answer. Results show that our approach makes an overall improvement with regard to Relatedness, Informativeness and Completeness and is less harmful to core fact retrieval. We also find that {\tt AMR-SG} (10/100) can make a further improvement compared to {\tt AMR-SG} (10/30) by including more facts to construct {\tt AMR-SG}. It demonstrates that {\tt AMR-SG} has the capability of detecting useful facts from a large and noisy fact pool, thus making up for the deficiency of the IR system.

\paragraph{Impact of AMR Consistency.}
We investigate the quality consistency of AMR graphs to see how it affects the construction of {\tt AMR-SG} and thus affects the QA model. We prepare AMR in three consistency levels, where \textit{Fully-Automatic} is generated by automatic AMR parser; \textit{Mixed} is that we manually annotate the error-free AMRs for the core facts in open-book (1326 in total) and use the error-free core fact AMRs and other automatically generated AMRs to construct {\tt AMR-SG}; \textit{Error-Free-Adapted} is that we use the error-free AMRs annotated to fine-tune the AMR parser and use the tuned parser to generate AMR for all the remaining facts (including hypotheses and common facts, about 900k in total). The test set accuracy are 81.6, 80.2, 80.4 for \textit{Fully-Automatic}, \textit{Mixed} and \textit{Error-Free-Adapted} respectively. It is interesting to note that using \textit{Fully-Automatic} AMRs results in higher QA accuracy than \textit{Mixed} and \textit{Error-Free-Adapted}, where the latter two contain a mix of AMRs with different levels of quality. This phenomenon has also been observed in other AMR applications~\cite{liu-etal-2015-toward, hardy-vlachos-2018-guided}, where automatic parses perform well than manual parses. We conjecture that this can be attributed to the discrepancy between the error-free AMRs and the automatically parsed AMRs in the choices of AMR concepts with similar meaning. This small difference in concept choices may omit potential connections, results in some important facts failing to be detected. In contrast, automatically parsed AMRs contain errors, but they are consistent in their concept choices, which is more likely for AMRs to form connections. The 0.2 accuracy improvement between \textit{Mixed} and \textit{Error-Free-Adapted} also demonstrates our assumption, since the parser is finetuned on the error-free AMRs, where its parsed AMRs should be more consistent with the error-free AMRs.
\begin{table}[]
    \centering
        \begin{tabular}{|p{73mm}|}
    \hline
         \textbf{Question:} \textit{A seismograph can accurately describe} (A) how rough the footing will be (B) how bad the weather will be \textbf{(C) how stable the ground will be} (D) how shaky the horse will be \\ \hline
         \textbf{Useful facts retrieved by IR:} N.A. \\ \hline
         \textbf{Additional facts from path-based analytics:} \\
         A seismograph is a kind of tool for measuring the size of an earthquake.
         
An earthquake is a shockwave travelling through the ground. \\ \hline
         \textbf{Relevant path in {\tt AMR-SG}}: \\
         {\tt seismograph$\rightarrow$tool$\rightarrow$measure-01$\rightarrow$ size-01$\rightarrow$earthquake$\rightarrow$ground
} \\
         \hline
    \end{tabular}
\caption{A case study showing how our framework selects useful facts to completely fill the knowledge gap.}
\label{case_study}
\end{table}
\begin{figure}[t]
    \centering
\includegraphics[scale=0.5]{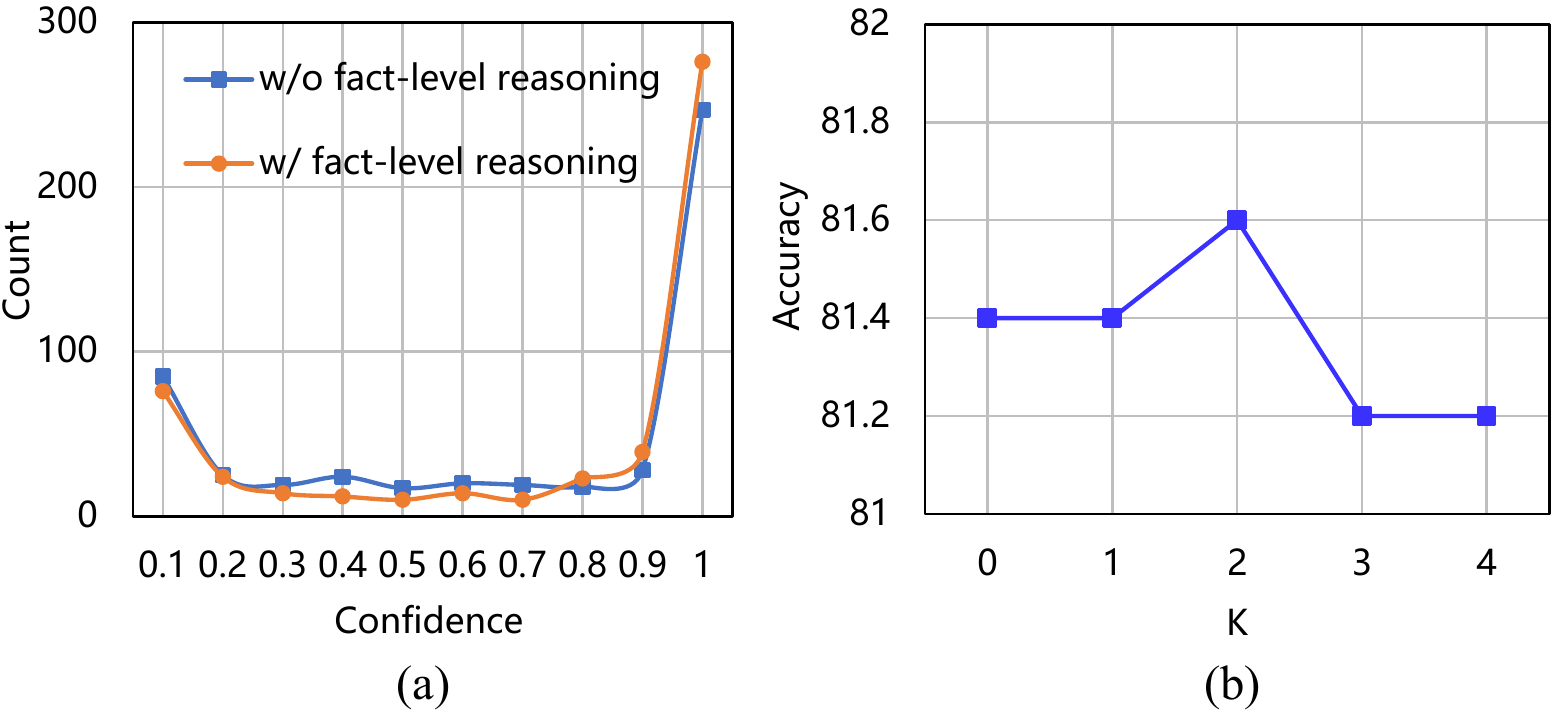} 
       \caption{Analysis of fact-level reasoning on OpenBookQA. (a) presents the distribution of prediction confidence with or without fact-level reasoning module. (b) shows the QA performance with different GCN layer K. Size 0 denotes the original pretrained model.}
    \label{fig:khop}
\end{figure}
\subsection{Case Study}
Table~\ref{case_study} shows one case study of evidence facts selected by our framework. Since the important term \textit{earthquake} is missing from the search query, the IR system assigns low retrieval scores for the two facts, causing a low ranking. However, the two facts can form a complete reasoning chain with the question and the answer via several concept nodes, where our approach can successfully extract the two facts despite the low retrieval scores. More cases can be found in Appendix~\ref{app:more_case}.

\subsection{Analysis of Fact-level Reasoning}
\paragraph{Why Fact-level reasoning.} Figure~\ref{fig:khop}(a) shows that fact-level reasoning improves the performance by making a more confident prediction for the correct answer. This is because the fact-level connections of {\tt AMR-SG} inform the model how these active facts are intrinsically related, which allows the model to precisely receive knowledge from related facts. 

\paragraph{Impact of Number of Hops (K).} We vary the hyper-parameter K to consider the impact of K-hop neighbors on OpenBookQA. As show in Figure~\ref{fig:khop}(b), the performance reaches the top at $K=2$. It indicates that most of the questions can be well answered using two evidence facts, which is consistent with the construction of this dataset. However, the performance drops when $K>2$. It might be attributed to exponential noise found in longer reasoning chains. 

\section{Conclusion}
We propose to dynamically construct {\tt AMR-SG} that can reflect the intrinsic relations of relevant facts leveraging AMR, a graph annotation. {\tt AMR-SG} combines the advantages of rich textual corpus and graph structure, where we can select useful facts that completely form the reasoning chain and make fact-level modeling. Experimental results show that {\tt AMR-SG} can maintain high explainability, and successfully couple with strong pretrained models to achieve significant improvement on OpenBookQA and ARC-Challenge over approaches leveraging additional KGs.

\bibliographystyle{acl_natbib}
\bibliography{acl2021}

\appendix
\section{Appendix}
\subsection{Case Study}
\label{app:more_case}
More case studies can be found in Table~\ref{tab:more_case}.
\begin{table*}[htbp]
    \centering
\subtable[Case Study 1]{
        \begin{tabular}{|p{47mm}|p{103mm}|}
    \hline
         \textbf{Question:} & \textit{Algae can be found in} \textbf{(A) reservoir} (B) meat (C) street (D) tree  \\ \hline
         \textbf{Useful facts retrieved by IR:} & Algae is found in bodies of water.
\\ \hline
         \textbf{Additional facts from path-based analytics:} & Water reservoir: a large quantity of water is stored in a reservoir (or dam).
\\ \hline
         \textbf{Relevant path in {\tt AMR-SG}}: & {\tt Algae $\rightarrow$ find-01 $\rightarrow$ body $\rightarrow$ water $\rightarrow$ store-01 $\rightarrow$ reservoir } \\
         \hline
    \end{tabular}
\label{3table}}
\subtable[Case Study 2]{
        \begin{tabular}{|p{47mm}|p{103mm}|}
    \hline
         \textbf{Question:} & \textit{Photosynthesis can be performed by} \textbf{(A) a cabbage cell} (B) a bee cell (C) a bear cell (D) a cat cell  \\ \hline
         \textbf{Useful facts retrieved by IR:} & N.A.\\ \hline
         \textbf{Additional facts from path-based analytics:} & Plant cells can perform photosynthesis.
         
        Description: skunk cabbage is a flowering perennial plant that is one of the first plants to emerge in the spring.\\ \hline
         \textbf{Relevant path in {\tt AMR-SG}}: & {\tt Photosynthesis $\rightarrow$ plant $\rightarrow$ cabbage} \\
         \hline
    \end{tabular}
\label{3table}}
\subtable[Case Study 3]{
    \begin{tabular}{|p{47mm}|p{103mm}|}
    \hline
         \textbf{Question:} &\textit{Which is recyclable?} (A) An Elephant \textbf{(B) A school notebook} (C) A boat (D) A lake  \\ \hline
         \textbf{Useful facts retrieved by IR:} & Paper is recyclable.\\ \hline
         \textbf{Additional facts from path-based analytics:} & Take notes on notebook paper. \\ \hline
         \textbf{Relevant path in {\tt AMR-SG}}: & {\tt recycle-01 $\rightarrow$ paper $\rightarrow$ notebook} \\
         \hline
    \end{tabular}
\label{firsttable}}
\subtable[Case Study 4]{
    \begin{tabular}{|p{47mm}|p{103mm}|}
    \hline
         \textbf{Question:} & \textit{Which requires energy to move?}
 \textbf{(A) weasel} (B) willow (C) mango (D) poison ivy  \\ \hline
         \textbf{Useful facts retrieved by IR:} & An animal requires energy to move.\\ \hline
         \textbf{Additional facts from path-based analytics:} & The long and slender body of the weasel allows it to move, almost flow, over terrain. \\ \hline
         \textbf{Relevant path in {\tt AMR-SG}}: & {\tt energy $\rightarrow$ move-01 $\rightarrow$ weasel} \\
         \hline
    \end{tabular}
\label{firsttable}}

\subtable[Case Study 5]{
    \begin{tabular}{|p{47mm}|p{103mm}|}
    \hline
         \textbf{Question:} &\textit{A person wants to be able to have more natural power in their home. They choose to cease using a traditional electric company to source this electricity, and so decide to install} (A) sun grafts (B) sunlight shields \textbf{(C) panels collecting sunlight} (D) solar bees  \\ \hline
         \textbf{Useful facts retrieved by IR:} & A home with solar electric panels on the roof might be able to make most of its own electricity, for example.\\ \hline
         \textbf{Additional facts from path-based analytics:} & Solar thermal panels generate hot water from the natural energy in sunlight. \\ \hline
         \textbf{Relevant path in {\tt AMR-SG}}: & {\tt natural-03 $\rightarrow$ energy $\rightarrow$ generate-01 $\rightarrow$ sunlight $\rightarrow$ panel} \\
         \hline
    \end{tabular}
\label{firsttable}}
    \caption{More case studies in addition to Table~\ref{case_study}}
    \label{tab:more_case}
\end{table*}

\end{document}